\documentclass{article}

\usepackage[preprint,nonatbib]{neurips_2026}

\usepackage[utf8]{inputenc} 
\usepackage[T1]{fontenc}    
\usepackage{hyperref}       
\usepackage{url}            
\usepackage{booktabs}       
\usepackage{amsfonts}       
\usepackage{mathtools}
\usepackage{xfrac}
\usepackage{nicefrac}       
\usepackage{microtype}      
\usepackage[table]{xcolor}         
\usepackage{graphicx}
\usepackage[sorting=none, backref=true]{biblatex}
\definecolor{linkcolor}{rgb}{0, 0.5, 128}
\hypersetup{colorlinks=true, citecolor=linkcolor, linkcolor=linkcolor, urlcolor=linkcolor}

\title{Flowing With Purpose: Latent Action Guided Flow Matching Policies For Robotic Manipulation}

\author{
  Bruno~Machado$^\spadesuit$ \quad Alexandre~Chapin \quad Emmanuel~Dellandréa \quad Liming~Chen \\
  École Centrale de Lyon \\
  LIRIS, UMR5205 \\
  $^\spadesuit$\texttt{bruno.machado-carneiro@ec-lyon.fr}
}

\begin{document}

\maketitle

\begin{abstract}
Flow matching has recently become a new standard for behavior cloning in robotic manipulation. However, state-of-the-art flow matching policies suffer from a systematic structural mismatch: they rely on a globally fixed isotropic source distribution despite the strongly fragmented and heteroscedastic structure of robotic action spaces. This agnostic initialization forces the model to learn highly entangled vector fields, bottlenecking training efficiency and limiting overall policy performance. To address this limitation, we introduce Latent Action Guided Flow Matching (LAFM), a novel framework that replaces the monolithic Gaussian with an adaptive library of learned prior distributions. By grounding these distributions using a latent action model, LAFM maps current observations to discrete motion primitives, selecting a specialized base distribution that provides an informed, structurally aligned initialization for the denoising process. This dynamic adaptivity naturally accommodates heteroscedasticity in human demonstrations and makes transport trajectories shorter and less entangled. Empirically, LAFM substantially outperforms standard flow matching formulations, increasing task success rates by 23.4\% in real-world robotic deployments and by 10.4\% on the LIBERO-90 benchmark. Furthermore, we demonstrate that LAFM achieves state-of-the-art results, surpassing massively pre-trained vision-language-action models while utilizing significantly smaller architectures. All code and model weights will be publicly released upon acceptance of the publication.
\end{abstract}

\section{Introduction}

Modeling visuomotor policy learning as a generative denoising process has recently emerged as a highly effective paradigm for robotic manipulation~\cite{diffusion-policy, pi0, gr00t}, with diffusion~\cite{ddpm, diffusion} and flow matching~\cite{flow-matching, rectified-flow} policies establishing new performance standards across both simulated and real-world benchmarks. Among these approaches, flow matching offers a particularly attractive formulation by directly learning a continuous transport field from a simple source distribution to the target action distribution, enabling faster inference and more stable optimization than diffusion-based alternatives.

Despite their empirical success, current flow matching policies rely on a strong structural assumption: all action trajectories are generated from a single isotropic Gaussian prior. While this assumption is convenient, it becomes restrictive in robotic manipulation, where the conditional action distribution is inherently fragmented and heterogeneous. Under nearly identical visual observations, multiple valid motor futures may coexist, corresponding to different grasp strategies, motion phases, or object interaction intents. As a result, the model must learn transport fields that connect a single unstructured source distribution to sharply separated action modes, forcing nearby noise samples to evolve toward distant targets and yielding unnecessarily entangled vector fields.

This limitation is further amplified by the heteroscedastic nature of manipulation demonstrations. Certain motion patterns, such as repetitive reaching or grasp closure, exhibit low variance and strong consistency, whereas others—particularly long-horizon object interactions—remain substantially more diverse. A single fixed prior cannot capture such mode-dependent uncertainty, implicitly imposing homogeneous variance across fundamentally different motion regimes.

These observations suggest that the source distribution itself should be structured before transport begins. Rather than initiating all denoising trajectories from the same global Gaussian, we argue that flow matching should start from priors already aligned with coarse motion intent. A suitable partition variable for this purpose must capture high-level dynamics rather than instantaneous controls. Latent action models (LAM)~\cite{genie}, recently introduced as unsupervised abstractions of temporal transitions, naturally provide such a representation by compressing visual dynamics into discrete motion primitives.

\begin{figure}[t]
    \centering
    \includegraphics[width=\linewidth]{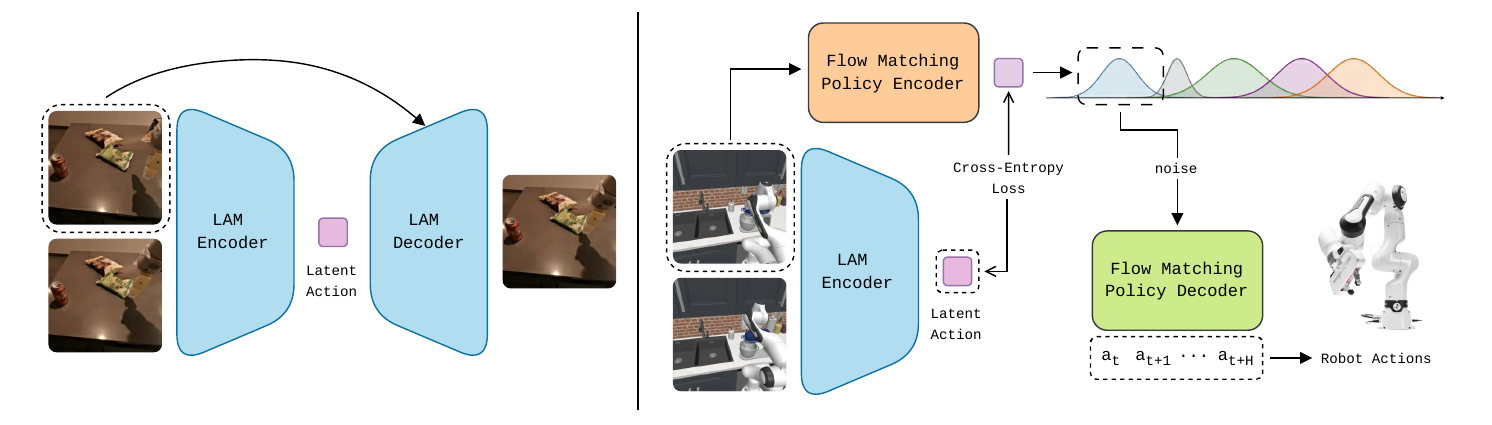}
    \caption{\textbf{LAFM overview.} Left: We train a LAM to extract latent actions from consecutive frames. Right: We divide our policy into two parts: The encoder predicts the next latent action for a given frame. The predicted latent action is used to condition the noise sampling, guiding the flow matching process in the decoder.}
    \label{fig:method}
\end{figure}

Building on this idea, we introduce Latent Action Guided Flow Matching (LAFM), a framework that replaces the monolithic Gaussian prior with a library of learned prior distributions indexed by latent actions. Instead of forcing the model to transport generic noise toward every possible action mode, LAFM first predicts a latent motion primitive from the current observation and then samples from a prior specialized for that primitive. This produces a more structured initialization for the generative process, yielding smoother transport trajectories and reducing vector field entanglement during training. An overview of the method is shown in Figure~\ref{fig:method}.

Importantly, we do not use latent actions simply as a pretraining signal, as done in recent latent-action-based policy learning approaches~\cite{lapa}. Instead, we use them directly as an explicit inductive bias for generative denoising policy learning.

Our work makes three main contributions:

\begin{itemize}
    \item We introduce LAFM, a new formulation of conditional flow matching in which the source distribution is no longer fixed but selected from a library of learnable priors. Each prior is associated with a latent motion primitive, allowing the generative process to begin from a distribution aligned with coarse action intent rather than from an uninformative global Gaussian.
    \item We propose a simple yet effective mechanism to ground these priors using a LAM. A policy encoder predicts discrete latent actions directly from current observations, and the predicted latent action determines which prior distribution parameterizes the noisy initialization of the decoder. This transforms latent actions from a pretraining signal into a direct structural bias on the geometry of the transport process.
    \item We show that structuring the source distribution substantially enhances final control performance in both simulation and real-world robotic manipulation settings. LAFM improves the success rate by 23.4\% in real-world experiments and 10.4\% on LIBERO-90, while outperforming significantly larger vision-language-action models (VLA).
\end{itemize}

\section{Methodology}

\subsection{Preliminary: the flow matching policy}

We follow the flow matching policy formulation proposed by \textcite{pi0}. Formally, the objective is to model the data distribution $p(\mathbf{A}_t|\mathbf{o}_t)$ of future actions $\mathbf{A}_t = [\mathbf{a}_t, \mathbf{a}_{t+1}, \dots, \mathbf{a}_{t+H-1}] $, conditioned on the current observation $\mathbf{o}_t$. The subscripts refer to robot control timesteps and $H$ denotes the action horizon. A neural network parametrized by $\theta$ is optimized to minimize the conditional flow matching objective~\cite{flow-matching, rectified-flow}:

\begin{equation} \label{eq:fm_loss}
    \mathcal{L}_{FM}^\tau (\theta) = \mathbb{E}_{p(\mathbf{A}_t|\mathbf{o}_t), q(\mathbf{A}_t^\tau|\mathbf{A}_t)}||\mathbf{v}_\theta(\mathbf{A}_t^\tau, \mathbf{o}_t) - \mathbf{u}(\mathbf{A}_t^\tau|\mathbf{A}_t)||^2
\end{equation}

The optimal transport probability path is defined as $q(\mathbf{A}_t^\tau | \mathbf{A}_t) = \mathcal{N}(\tau \mathbf{A}_t, (1 - \tau)^2\mathbf{I})$, where $\tau \in [0, 1]$ represents the flow matching step. This path is realized by sampling standard Gaussian noise $\epsilon \sim \mathcal{N}(\mathbf{0}, \mathbf{I})$ to construct noisy action targets through linear interpolation: $\textbf{A}_t^\tau = \tau \mathbf{A}_t + (1 - \tau)\epsilon$. The flow matching step $\tau$ is sampled from a Beta distribution that prioritizes noisier samples (see Appendix~\ref{appendix:policies}). The network output $\mathbf{v}_\theta(\mathbf{A}_t^\tau, \mathbf{o}_t)$ is explicitly optimized to regress the denoising vector field $\mathbf{u}(\mathbf{A}_t^\tau | \mathbf{A}_t) = \mathbf{A}_t - \epsilon$.

During inference, actions are predicted by integrating the vector field from $\tau=0$ to $\tau=1$. This is implemented using Euler integration rule: $\mathbf{A}_t^{\tau+\delta} = \mathbf{A}_t^\tau + \delta \mathbf{v}_\theta(\mathbf{A}_t^\tau, \mathbf{o}_t)$, where $\delta$ is the integration step size. Starting with $\mathbf{A}_t^0 = \epsilon$, \sfrac{1}{$\delta$} integration steps are performed to obtain the final prediction $\mathbf{A}_t^1$.

\subsection{Adaptive flow matching with multiple priors}
\label{sec:adafm}

Instead of using a single prior $\mathcal{N}(\mathbf{0}, \mathbf{I})$, we propose using a library of $K$ learned prior distributions $\{\mathcal{N}(\boldsymbol{\mu}_k, \boldsymbol{\Sigma}_k)\}_{k=1}^K$. For each data sample, the network predicts a latent mode distribution $\pi(\mathbf{o}_t)$ conditioned on the current observation, where $\pi_k(\mathbf{o}_t)$ denotes the probability that the action belongs to latent mode $k$. An index $k$ is then selected, and noise is sampled from the corresponding distribution: $\epsilon_k \sim \mathcal{N}(\boldsymbol{\mu}_k, \boldsymbol{\Sigma}_k)$. Because both the prior parameters and the selection mechanism are learned, the resulting probability path is adaptive and parameter-dependent: $q_\theta(\mathbf{A}_t^\tau | \mathbf{A}_t) = \mathcal{N}\left( \tau \mathbf{A}_t + (1-\tau) \boldsymbol{\mu}_{k}, (1-\tau)^2 \boldsymbol{\Sigma}_{k} \right)$. As shown in Equation~(\ref{eq:adafm_loss}), the objective is to regress the adaptive denoising vector field $\mathbf{u}_\theta(\mathbf{A}_t^\tau | \mathbf{A}_t) = \mathbf{A}_t - \epsilon_k$.

\begin{equation} \label{eq:adafm_loss}
    \mathcal{L}_{AFM}^\tau (\theta) = \mathbb{E}_{p(\mathbf{A}_t|\mathbf{o}_t), q_\theta(\mathbf{A}_t^\tau|\mathbf{A}_t)}||\mathbf{v}_\theta(\mathbf{A}_t^\tau, \mathbf{o}_t) - \mathbf{u}_\theta(\mathbf{A}_t^\tau|\mathbf{A}_t)||^2
\end{equation}

This structured source distribution admits a simple geometric interpretation: when each latent prior aligns with a distinct mode of the conditional action distribution, the expected transport burden imposed on the learned vector field is reduced. We formalize this intuition with Proposition~\ref{prop:1}.

\newtheorem{prop}{Proposition}
\begin{prop}\label{prop:1}\textbf{Expected transport reduction under latent-conditioned priors}

Let $p(\mathbf{A}_t|\mathbf{o}_t)$ denote a conditional action distribution decomposable into $K$ latent motion modes: $p(\mathbf{A}_t|\mathbf{o}_t)=\sum_{k=1}^{K}\pi_k(\mathbf{o}_t)p_k(\mathbf{A}_t|\mathbf{o}_t)$, where each component $p_k(\mathbf{A}_t|\mathbf{o}_t)$ has mean $\mathbf{m}_k(\mathbf{o}_t)$. Consider two source constructions: a single isotropic prior ($\epsilon \sim \mathcal{N}(\mathbf{0}, \mathbf{I})$) and a latent-conditioned prior ($\epsilon_k \sim \mathcal{N}(\boldsymbol{\mu}_k, \boldsymbol{\Sigma}_k)$) selected according to latent mode $k$. If each prior satisfies $\boldsymbol{\mu}_k~\approx~\mathbb{E}_{\mathbf{o}_t}[\mathbf{m}_k(\mathbf{o}_t)]$ and $\mathrm{Tr}(\boldsymbol{\Sigma}_k) \leq \mathrm{Tr}(\mathbf{I})$, then the expected squared transport distance satisfies $\mathbb{E}||\mathbf{A}_t-~\epsilon_k||^2~\leq~\mathbb{E}||\mathbf{A}_t~-~\epsilon||^2$ under perfect mode coupling. 
\end{prop}

In practice, exact alignment $\boldsymbol{\mu}_k~=~\mathbb{E}_{\mathbf{o}_t}[\mathbf{m}_k(\mathbf{o}_t)]$ is not required. Latent supervision only needs to induce a coarse partition of the action space such that samples within each partition are more transport-aligned than samples across partitions. A proof sketch for Proposition \ref{prop:1} is provided in Appendix~\ref{appendix:proofs}.

Importantly, the proposed priors are not used as output mixture components, but as source distributions defining distinct transport paths during training and inference. Unlike mixture-of-Gaussians conditioning, our method modifies the generative trajectory itself rather than only increasing output expressiveness.

\subsection{LAFM}

To ensure the prior distributions are informative and provide a structured initialization for the denoising process, we leverage a LAM, which is pre-trained to compress the dynamics between two temporal observations, $\mathbf{o}_t$ and $\mathbf{o}_{t+T}$, into a discrete latent space $\mathcal{C} = \{1, \dots, C\}$. We set the number of prior distributions $K$ equal to the codebook size $C$, establishing a direct correspondence between each discrete latent action $\ell_a \in \mathcal{C}$ and a specific prior $\mathcal{N}(\boldsymbol{\mu}_k, \mathbf{\Sigma}_k)$. The policy network predicts a categorical distribution $P_{\theta_{cls}}(k \mid \mathbf{o}_t) \equiv \pi_k(\mathbf{o}_t)$, which serves as the latent mode distribution defined in Section~\ref{sec:adafm}. The model is trained to predict the latent action index $\ell_a \coloneqq \phi_e(\mathbf{o}_t, \mathbf{o}_{t+T})$, where $\phi_e$ represents the pre-trained LAM encoder. The predicted distribution $P_{\theta_{cls}}(k | \mathbf{o}_t)$ is supervised using the discrete latent action $\ell_a$ via a categorical cross-entropy (CCE) loss:

\begin{equation} \label{eq:cce_loss}
    \mathcal{L}_{CCE}(\theta_{cls}) = -\mathbb{E}_{(\mathbf{o}_t, \ell_a)} \left[ \log P_{\theta_{cls}}(k = \ell_a | \mathbf{o}_t) \right]
\end{equation}

where $\theta_{cls} \subset \theta$ . The predicted index $k$ is selected with an operation $\arg\max$, and the corresponding prior is used for flow matching. In this way, the network learns to bias the denoising process toward the specific motion primitive identified by the latent action.

All learned prior distributions are initialized as standard normal distributions, $\mathcal{N}(\mathbf{0}, \mathbf{I})$. This initialization strategy is crucial for training stability; early in the training process, the model's latent action predictions are largely uninformative, and a uniform starting point prevents the flow matching objective from diverging. As training progresses, the priors naturally specialize, adapting their means and covariances to capture the specific spatial regions of their corresponding motion primitives. However, over-specialized priors could lead to overfitting, particularly for underrepresented motion primitives. To prevent this, we introduce a Kullback-Leibler (KL) divergence regularization term. This auxiliary loss ensures that the learned priors do not deviate excessively from a manageable search space, by penalizing the divergence between the selected prior distribution and a standard normal distribution. The KL divergence term is computed analytically in closed-form as:

\begin{equation} \label{eq:kl_analytical}
    \mathcal{R}_{KL}(\theta) = \mathbb{E}_{k \sim P_{\theta_{cls}}(k | \mathbf{o}_t)} \left[ \frac{1}{2} \sum_{d=1}^D \left( \boldsymbol{\mu}_{k, d}^2 + \boldsymbol{\sigma}_{k, d}^2 - 1 - 2\log\boldsymbol{\sigma}_{k, d} \right) \right]
\end{equation}

The overall objective for training a model with LAFM integrates the three aforementioned components:

\begin{equation} \label{eq:final_loss}
    \mathcal{L}_{LAFM}(\theta) = \mathcal{L}_{AFM}^\tau(\theta) + \lambda_1 \mathcal{L}_{CCE}(\theta_{cls}) + \lambda_2 \mathcal{R}_{KL}(\theta)
\end{equation}

where $\lambda_1$ and $\lambda_2$ are scalar weighting hyperparameters. The KL regularization weight $\lambda_2$ controls the degree of deviation of the learned priors from the base isotropic distribution. In particular, as $\lambda_2 \to \infty$, minimizing $\mathcal{R}_{KL}$ enforces $\boldsymbol{\mu}_k \to \mathbf{0}$ and $\boldsymbol{\Sigma}_k \to \mathbf{I}$ for all $k$. In this limit, the adaptive probability path becomes indistinguishable from the standard flow matching path, making the optimization virtually equivalent to solving the classical flow matching objective (Equation~(\ref{eq:fm_loss})) alongside the latent classification task (Equation~(\ref{eq:cce_loss})). Conversely, lowering $\lambda_2$ relaxes this constraint, granting the network the flexibility to learn highly specialized, localized prior distributions. Thus, tuning $\lambda_2$ provides a continuous interpolation between a constrained, single-prior baseline and a fully adaptive, multi-prior ensemble. 
   
\subsection{Implementation details} \label{sec:arch}

\begin{figure}[h]
    \centering
    \includegraphics[width=\linewidth]{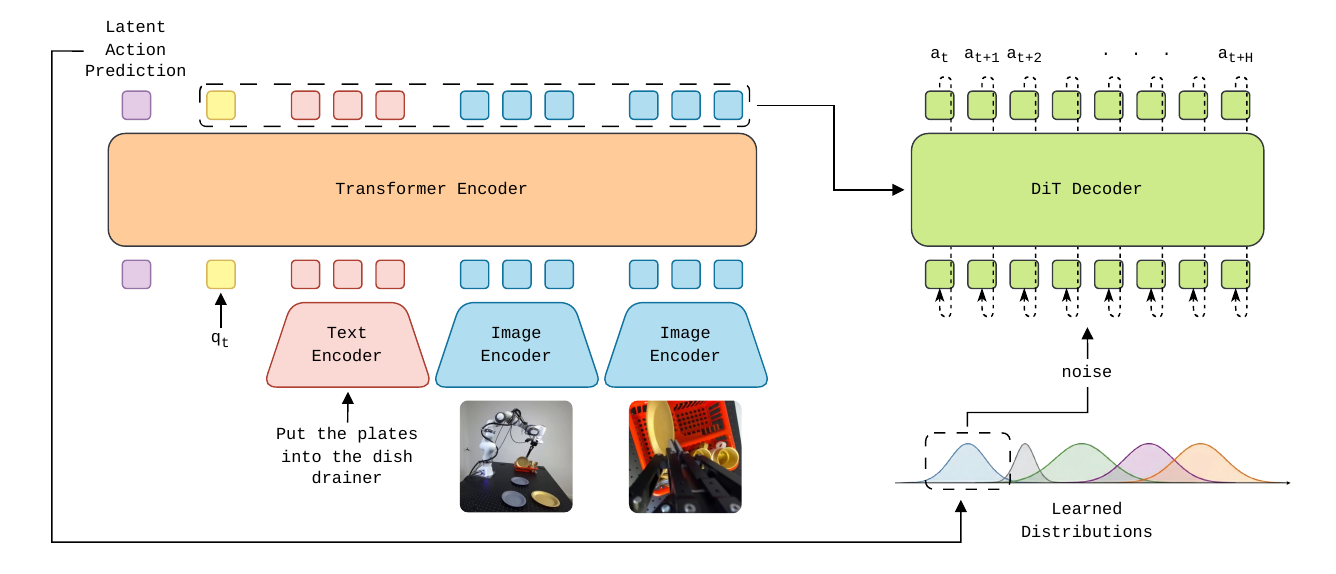}
    \caption{\textbf{Implemented architecture.} The images, task description, and robot's proprioception are given to the policy encoder. The images and text are first processed by a specialized encoder before being fed to a Transformer. The output of the Transformer encoder is given via cross-attention to the DiT decoder. The latent action predicted by the encoder is used to choose the distribution from which the noise is sampled.}
    \label{fig:arch}
\end{figure}

In order to validate our method, we propose an encoder-decoder architecture to implement the flow matching policy, as depicted in Figure~\ref{fig:arch}. For all of our experiments, we consider an observation $\mathbf{o}_t = [\mathbf{I}_t^1, ..., \mathbf{I}_t^n, l_t, \mathbf{q}_t]$, which incorporates one or more RGB images ($\mathbf{I}_t^1, ..., \mathbf{I}_t^n$), a description of the target task given in natural language ($l_t$), and the proprioceptive state of the robot ($\mathbf{q}_t$). Task instructions are embedded via a frozen text encoder, whereas images are processed by a pre-trained image encoder, which is further optimized during training. These preprocessed features, together with the proprioceptive state vector, are projected to a unified dimension, forming a multi-modal token sequence. An additional dedicated prompt token is added to the beginning of the sequence, which is processed by a randomly initialized Transformer~\cite{attention} using full bidirectional attention~\cite{bert}. 

The hidden state corresponding to the prompt token is linearly projected to the codebook dimension $C$ to produce classification logits (Equation~(\ref{eq:cce_loss})). The predicted index $k$ acts as a lookup key for the prior library, which is implemented using two embedding layers: $\mathbf{E}_{\boldsymbol{\mu}}, \mathbf{E}_{ \boldsymbol{\log \sigma}} \in \mathbb{R}^{C \times D}$, where $C$ is the number of priors and $D$ is the dimensionality of the action space. These embeddings store the learnable parameters $\boldsymbol{\mu}_k$ and $\boldsymbol{ \log\sigma}_k$ for each motion primitive. To sample the contextual noise $\epsilon_{k}$, we first draw a sample from a standard normal distribution $\zeta \sim \mathcal{N}(\mathbf{0}, \mathbf{I})$ and then apply a location-scale transformation using the retrieved embedding vectors: $\epsilon_{k} = \boldsymbol{\mu}_{k} + \boldsymbol{\sigma}_{k} \odot \zeta$, where $\boldsymbol{\sigma}_{k} = \exp(\boldsymbol{\log \sigma}_{k})$ and $\odot$ denotes the element-wise product.

The decoder is parameterized as a Diffusion Transformer (DiT)~\cite{dit}, which is conditioned on the output of the encoder through cross-attention. The input to the decoder consists of the noisy action sequence $\mathbf{A}_t^\tau$. Consequently, the input sequence comprises $H$ tokens, which are linearly projected from the original action space to the hidden dimension of the decoder. Analogous to the encoder architecture, we do not employ attention masking. The model is optimized using the LAFM loss formulated in Equation~(\ref{eq:final_loss}). 

\paragraph{LAM implementation details.} 
A comprehensive description of the deployed LAM is provided in Appendix~\ref{appendix:lam}, while Appendix~\ref{appendix:lam-ablations} presents ablation studies investigating the impact of LAM training data and codebook size on policy performance.

\section{Experiments}
\label{sec:exp}

We evaluate our method in both real-world robotic manipulation settings and simulated environments. We design experiments to address the following questions:

\begin{itemize}
    \item Does our method improve the standard flow matching policy formulation and is the improvement significant?
    \item How does our method compare with the state of the art? Can it be competitive with policies pre-trained on large-scale robotic datasets?
    \item Are multiple priors necessary, or is latent action prediction supervision sufficient?
\end{itemize}

\subsection{Main results}

To evaluate our proposed method, we establish a direct comparison with a standard flow matching baseline (FM). To ensure a fair evaluation, the FM model shares the exact architecture and hyperparameters of LAFM, as described in Section~\ref{sec:arch}, but omits the latent action prediction module. Instead, it relies on sampling noise from a single fixed normal distribution, being optimized via the standard conditional flow matching loss (Equation~(\ref{eq:fm_loss})). Additionally, to contextualize our performance within the broader landscape of manipulation policies, we benchmark LAFM against a comprehensive suite of state of the art methods~\cite{diffusion-policy, octo, openvla, pi0, smolvla, univla, act, vqbet, quest}.

\subsubsection{Real-world experiments}

We evaluate our policies on four distinct manipulation tasks, using 50 demonstrations per task. These demonstrations are collected with a GELLO teleoperation system~\cite{gello} to control a Franka Emika Panda robotic arm. The tasks consist of: (1) picking up all \textbf{plates} from the workspace and placing them in a dish drying rack; (2) opening a \textbf{drawer}, picking up a screwdriver, stowing it inside, and closing the drawer; (3) picking up all \textbf{cans} from the table and depositing them into a trash bin; and (4) stacking all \textbf{bowls} on the table into a single pile. Figure~\ref{fig:tasks} provides a visualization of each task.

Due to the prohibitive time and resource costs associated with real-world training and evaluation, we restrict our comparative analysis to two primary baselines. First, we consider ACT~\cite{act} as a comparably sized model which is trained from randomly initialized weights. Note that while LAFM leverages a pretrained LAM for dataset labeling, the policy model is also learned from scratch. Second, we include $\pi_0$~\cite{pi0} to assess how our approach compares to a VLA with substantially more parameters and extensive policy pre-training. Detailed training configurations for each policy are provided in Appendix~\ref{appendix:policies}.

We report two primary metrics for our real-world evaluations: success rate and completion score. Success rate serves as a strict binary criterion for task completion. In contrast, the completion score is a continuous metric ranging from 0 to 1 that indicates the fraction of the task successfully executed. Comprehensive task descriptions and the precise criteria used to compute completion scores are detailed in Appendix~\ref{appendix:experiments}. We conduct 15 evaluation trials per task for each model, randomizing object positions for every trial.

\begin{table}[h!]
    \caption{\textbf{Results on real-world.} Success rates (\%) and completion score (CS) (\%) for various policies. VLA Pt refers to policy pretraining on robotics data.}
    \centering
    \setlength{\tabcolsep}{6pt}
    \renewcommand{\arraystretch}{1.2}
    \definecolor{grayrow}{gray}{0.90}
    \definecolor{graycol}{gray}{0.95}
    \newcolumntype{g}{>{\columncolor{graycol}}c}
    \begin{tabular}{lcccccgg}
        \toprule
        \textbf{Policy (\# Params)} & \textbf{VLA Pt.} & \multicolumn{5}{c}{\textbf{Success Rate (\%)}} & \multicolumn{1}{c}{\textbf{CS (\%)}}
             \\
        \midrule
        \rowcolor{grayrow}
        \multicolumn{2}{l}{} & \textbf{Plates} & \textbf{Drawer} & \textbf{Cans} & \textbf{Bowls} & \textbf{Avg.} & \textbf{Avg.}\\
        $\pi_0$ (3.3B) \cite{pi0} & Yes & 86.7 & 60.0 & 66.7 & 73.3 & 71.7 & 82.9 \\
        ACT (0.05B) \cite{act} & No & 40.0 & 13.3 & 60.0 & 46.7 & 40.0 & 51.7  \\
        \midrule
        FM (0.11B) & No & 73.3 & 46.7 & 66.7 & 66.7 & 63.3 & 75.0 \\
        LAFM (0.11B) & No & \textbf{100.0} & \textbf{80.0} & \textbf{86.7} & \textbf{80.0} & \textbf{86.7} & \textbf{93.9} \\
        \bottomrule
    \end{tabular}
    \label{tab:real_world}
\end{table}

Our method improves the average success rate by 23.4\% over FM and 15.0\% over $\pi_0$, with corresponding completion score gains of 18.9\% and 11.0\%, respectively. Qualitatively, our policy rollouts exhibit much smoother, more fluid motions compared to FM.

\subsubsection{LIBERO}

To ease reproducibility and provide a rigorous comparison against diverse baselines, we evaluate our method on LIBERO, a simulated robotic manipulation environment comprising five benchmark suites: Goal, Object, Spatial, Long (also referred to as LIBERO-10), and LIBERO-90. While several recent works~\cite{openvla, smolvla, univla} restrict their evaluations to the first four benchmarks, we advocate for a comprehensive evaluation across all five. However, to ensure a fair and direct comparison with these aforementioned works, we partition our empirical study into two phases. First, following \textcite{smolvla}, we train a single unified model across the four combined benchmarks (Spatial, Object, Goal, and Long). Subsequently, we conduct an independent evaluation solely on LIBERO-90.

\paragraph{LIBERO Spatial, Object, Goal and Long.}

Each of the four benchmarks comprises 10 tasks. We train all models using a preprocessed version of the dataset introduced by \textcite{openvla}, which yields a refined set of 1,693 episodes encompassing all 40 tasks. For all LIBERO evaluations, we execute 50 trials per task and average the results across 3 random seeds. We report the standard deviation with Bessel’s correction. Baseline metrics are sourced directly from \textcite{openvla,smolvla,univla}, with the exception of ACT, which we independently train and evaluate.

The evaluation results are detailed in Table~\ref{tab:libero}. It is important to note that Diffusion Policy~(DP)~\cite{diffusion-policy}, Octo~\cite{octo}, OpenVLA~\cite{openvla}, and UniVLA~\cite{univla} train benchmark-specific models, resulting in a distinct set of weights for each of the four suites. Consequently, the reported averages for these baselines reflect the mean performance of specialized models rather than a single, unified policy. Conversely, SmolVLA~\cite{smolvla} and $\pi_0$ train a single model concurrently on all four benchmarks. We adopt this unified training paradigm for all models trained in our study.

\begin{table}[h!]
    \caption{\textbf{Results on LIBERO Spatial, Object, Goal, and Long.} Success rates (\%) for various policies. VLA Pt refers to policy pretraining on robotics data. Baseline scores from \textcite{openvla,smolvla,univla}. $\pm$ denotes standard deviation. $\dagger$ denotes average performance of four distinct expert policies.}
    \centering
    \setlength{\tabcolsep}{3pt}
    \renewcommand{\arraystretch}{1.2}
    \definecolor{grayrow}{gray}{0.90}
    \definecolor{graycol}{gray}{0.95}
    \newcolumntype{g}{>{\columncolor{graycol}}c}
    \begin{tabular}{lcccccg}
        \toprule
        \textbf{Policy (\# Params)} & \textbf{VLA Pt.} & \multicolumn{5}{c}{\textbf{Success Rate (\%)}} \\
        \midrule
        \rowcolor{grayrow}
        \multicolumn{2}{l}{} & \textbf{Spatial} & \textbf{Object} & \textbf{Goal} & \textbf{Long} & \textbf{Avg.}\\
        DP (0.26B) \cite{diffusion-policy}  & No & 78.3 & 92.5 & 68.3 & 50.5 & 72.4$^{\dagger}$ \\
        Octo (0.09B) \cite{octo} & Yes & 78.9 & 85.7 & 84.6 & 51.1 & 75.1$^{\dagger}$ \\
        OpenVLA (7B) \cite{openvla} & Yes & 84.7 & 88.4 & 79.2 & 53.7 & 76.5$^{\dagger}$ \\
        $\pi_0$ (3.3B) \cite{pi0} & Yes & 90 & 86 & 95 & 73 & 86.0 \\
        SmolVLA (2.25B) \cite{smolvla} & No & 93	& 94 & 91 & 77 & 88.8 \\
        UniVLA (7B) \cite{univla} & Yes & 96.5	& 96.8 & 95.6 & 92.0 & 95.2$^{\dagger}$ \\
        ACT (0.05B) \cite{act} & No & $96.3 \pm 0.5$ & $\mathbf{99.5} \pm 0.1$ & $97.0 \pm 0.7$ & $90.8 \pm 0.9$ & $95.9 \pm 0.5$ \\
        \midrule
        FM (0.11B) & No & $96.1 \pm 1.0$ & $97.9 \pm 0.3$ & $96.5 \pm 0.5$ & $91.2 \pm 0.2$ & $95.4 \pm 0.3$ \\
        LAFM (0.11B) & No & $\mathbf{97.2} \pm 0.7$ & $99.2 \pm 0.0$ & $\mathbf{97.2} \pm 0.5$ & $\mathbf{92.3} \pm 0.3$ & $\mathbf{96.5} \pm 0.3$ \\
        \bottomrule
    \end{tabular}
    \label{tab:libero}
\end{table}

Empirically, FM demonstrates strong baseline performance, remaining highly competitive with significantly larger models that leverage large-scale pre-training. By incorporating our method, we achieve a consistent, albeit marginal, performance gain of approximately 1\% in success rate across all benchmarks, allowing our model to surpass all baselines. While this improvement is systematic across settings, the near-ceiling success rate of FM intrinsically limits the margin for further performance gains.

\paragraph{LIBERO-90.}

The LIBERO-90 suite comprises 90 distinct manipulation tasks. Applying the same data preprocessing protocol introduced by \textcite{openvla}, we filter the original dataset to obtain a final corpus of 3,959 episodes encompassing all 90 tasks. Baseline results are sourced from \textcite{quest}, with the exception of ACT and $\pi_0$, which we train and evaluate ourselves.

\begin{table}[h!]
    \caption{\textbf{Results on LIBERO 90.} Success rates (\%) for various policies. VLA Pt refers to policy pretraining on robotics data. Baseline scores from \textcite{quest}. $\pm$ denotes standard deviation.}
    \centering
    \setlength{\tabcolsep}{6pt}
    \renewcommand{\arraystretch}{1.2}
    \definecolor{grayrow}{gray}{0.90}
    \begin{tabular}{lcc}
        \toprule
        \textbf{Policy (\# Params)} & \textbf{VLA Pt.} & \textbf{Success Rate (\%)} 
             \\
        \midrule
        DP (0.26B) \cite{diffusion-policy} & No & 75.4 \\
        VQ-BeT \cite{vqbet} & No & 81.3 \\
        QueST \cite{quest} & No & 88.6 \\
        $\pi_0$ (3.3B) \cite{pi0} & Yes & $92.1 \pm 0.1$ \\
        ACT (0.05B) \cite{act} & No & $86.2 \pm 0.2$ \\
        \midrule
        FM (0.11B) & No & $82.6 \pm 0.2$ \\
        FM + LAC (0.11B) & No & $85.4 \pm 0.1$ \\
        LAFM (0.11B) & No & $\mathbf{93.0} \pm 0.2$ \\
        \bottomrule
    \end{tabular}
    \label{tab:libero_90}
\end{table}

This experiment more clearly highlights the efficacy of our approach, yielding an absolute gain of 10.4\% in success rate over FM. Furthermore, our method systematically outperforms all considered baselines, establishing state-of-the-art results on the LIBERO-90 benchmark.

\subsection{Impact of auxiliary supervision}

To isolate the efficacy of source-distribution structuring from the benefits of auxiliary representation learning, we compare our proposed LAFM against a standard flow matching policy augmented with an identical latent action prediction objective. Specifically, we evaluate a variant denoted as FM~+~LAC (Latent Action Conditioning). In this setup, the model continues to sample from a single standard Gaussian prior, and the predicted latent action is explicitly provided to the decoder via cross-attention.

We conduct this experiment on LIBERO-90, with results reported in Table~\ref{tab:libero_90}. Adding latent action prediction as an auxiliary supervision improves the performance of the standard flow matching policy. However, this improvement does not fully account for the gains achieved by LAFM, highlighting the importance of structured priors in our approach.

\subsection{Qualitative analysis}

\subsubsection{Learned priors analysis}

Figure~\ref{fig:priors} shows a t-SNE visualization of the learned prior distributions for our model trained on LIBERO. We display the projected means, representing the variance by color and size of the circles. To better understand the correspondence of the learned distributions and the latent actions, we take a single image as input and generate an output image using each possible latent action and feeding them to the LAM's decoder. We then compute the optical flow between the generated images and the original image. We display the original image with a superposed optical flow representation for a couple of latent actions that have similar learned priors. 

We can observe that our model learns similar distributions for latent actions that represent similar movements. Another interesting thing to note is that latent actions which represent larger range movements have corresponding priors with larger variances, while smaller range movements have more specialized priors, with smaller variance. This reinforces our intuition that latent actions are suitable to capture the heteroscedasticity of human demonstrations, making them appropriate for guiding our priors.

\subsubsection{Vector fields visualization}
\label{sec:vector-vis}

To have an intuition of how our method simplifies flow matching, we perform the inference for our model and FM for the same random batch of 512 items from LIBERO-90. We use $\delta=0.1$ and store the intermediate predictions for each denoising step. We use PCA to reduce the dimension of the action space, computing it based on the target actions. We then project the original noise points and the model's predictions into the first and second principal components, interpolating the 9 intermediate predictions to represent the denoising vector field. To allow for a visualization of the transport process, we shift the mean of the prior distributions by adding an offset to all the noise points, moving them to the left. For the outputs of the model, the offset is proportional to $1 - \tau$, being zero for the final prediction. 

The resulting graphs are shown in Figure~\ref{fig:vector-fields}, which empirically illustrates Proposition \ref{prop:1}: latent-conditioned priors reduce the initial displacement between source and target samples, producing shorter and less entangled transport trajectories, considerably reducing the transport cost and the final prediction error.

\section{Related works}

\subsection{Diffusion and flow matching policies}

Diffusion Policy \cite{diffusion-policy} introduced the concept of modeling a robot's policy as a conditional denoising diffusion process, quickly establishing itself as a standard approach for action prediction in robotic manipulation \cite{octo}. Building upon this foundation, $\pi_0$ \cite{pi0} replaced the diffusion objective with a flow matching formulation—a shift adopted by subsequent research \cite{gr00t, } that has since emerged as the new state of the art. While diffusion models generate data by learning to reverse a gradual, stochastic noise-adding process step-by-step, flow matching achieves this by directly learning a continuous vector field that transports a simple base distribution straight to a complex target distribution, allowing faster inference using fewer denoising steps. Beyond architectural differences, our work extends flow matching policies by learning distinct prior distributions, whereas these previous methods sample noise from a single, fixed distribution.

\subsection{Noise optimization for flow matching policies}

DSRL \cite{dsrl} uses Reinforcement Learning (RL) to optimize the noise space of pre-trained diffusion and flow matching policies, adapting them to downstream scenarios without altering the models' weights. Parrot \cite{parrot} employs a similar strategy, but uses a policy based on normalizing flow. Similarly, we also adjust the prior noise distribution of the policy to improve its performance. But unlike these methods, which use RL to adapt trained policies, our approach operates directly during policy training, learning different noise distributions based on the bias introduced by a LAM.

\subsection{Latent action models applied to robotics}

Latent action models (LAM) function as unlabeled world models that learn directly from video, without requiring ground-truth action labels. Genie \cite{genie} uses a LAM to label videos, mapping the learned latent actions to user inputs to create interactive environments. In the context of 2D video games, ILPO \cite{ilpo} and LAPO \cite{lapo} learn a policy within the latent action space before mapping those latent actions to real actions. LAPA \cite{lapa} utilizes latent actions to pre-train a VLA, which establishes a unified action space across different embodiments and allows the model to leverage unlabeled videos (i.e., human manipulation videos) as part of its pre-training dataset. Subsequent studies have also adopted latent actions as a labeling technique for VLA pre-training, using them either as the sole pre-training targets \cite{moto} \cite{univla} or alongside other training objectives \cite{gr00t}. Unlike prior latent action based policy learning methods, which use latent actions simply as pretraining supervision targets, our method leverages latent actions as a guiding bias to generative policy learning, using them to directly reshape the source distribution, thereby altering the geometry of the learned transport field.

\section{Conclusion}

In this work, we introduced LAFM, a novel generative framework designed to overcome the structural inefficiencies inherent in standard flow matching policies for robotic manipulation. By transitioning from a monolithic Gaussian prior to an adaptive library of learned distributions grounded by a LAM, our approach provides a highly informative initialization for the denoising process. This structurally informed starting point leads to more direct transport trajectories, disentangles the underlying vector fields, and naturally accommodates the heteroscedasticity of multi-modal human demonstration data. 

Our extensive empirical evaluations demonstrate that LAFM yields substantial improvements over standard flow matching formulations. In real-world robotic deployments, our method increased the task success rate by 23.4\%, and by 10.4\% on the LIBERO-90 benchmark. Crucially, we showed that LAFM achieves state-of-the-art performance, surpassing significantly larger VLA baselines, such as $\pi_0$ and UniVLA. 

Despite these strong results, our framework presents distinct avenues for future exploration. Currently, our reliance on video-based LAMs means that the latent representations can occasionally capture extraneous visual dynamics, such as moving backgrounds or severe camera shifts, rather than strictly isolating the robot's kinematics. Future work will focus on refining the information bottleneck to better disentangle agent-specific movements from environmental noise. Additionally, integrating our multi-prior flow matching formulation directly into the large-scale pre-training pipelines of foundational VLA architectures represents a highly promising next step. Ultimately, LAFM establishes a robust, efficient, and scalable paradigm for teaching robots complex behaviors through structured generative modeling.

\printbibliography

\newpage

\appendix

\section{Qualitative analysis figures}

\begin{figure}[h!]
    \centering
    \includegraphics[width=\linewidth]{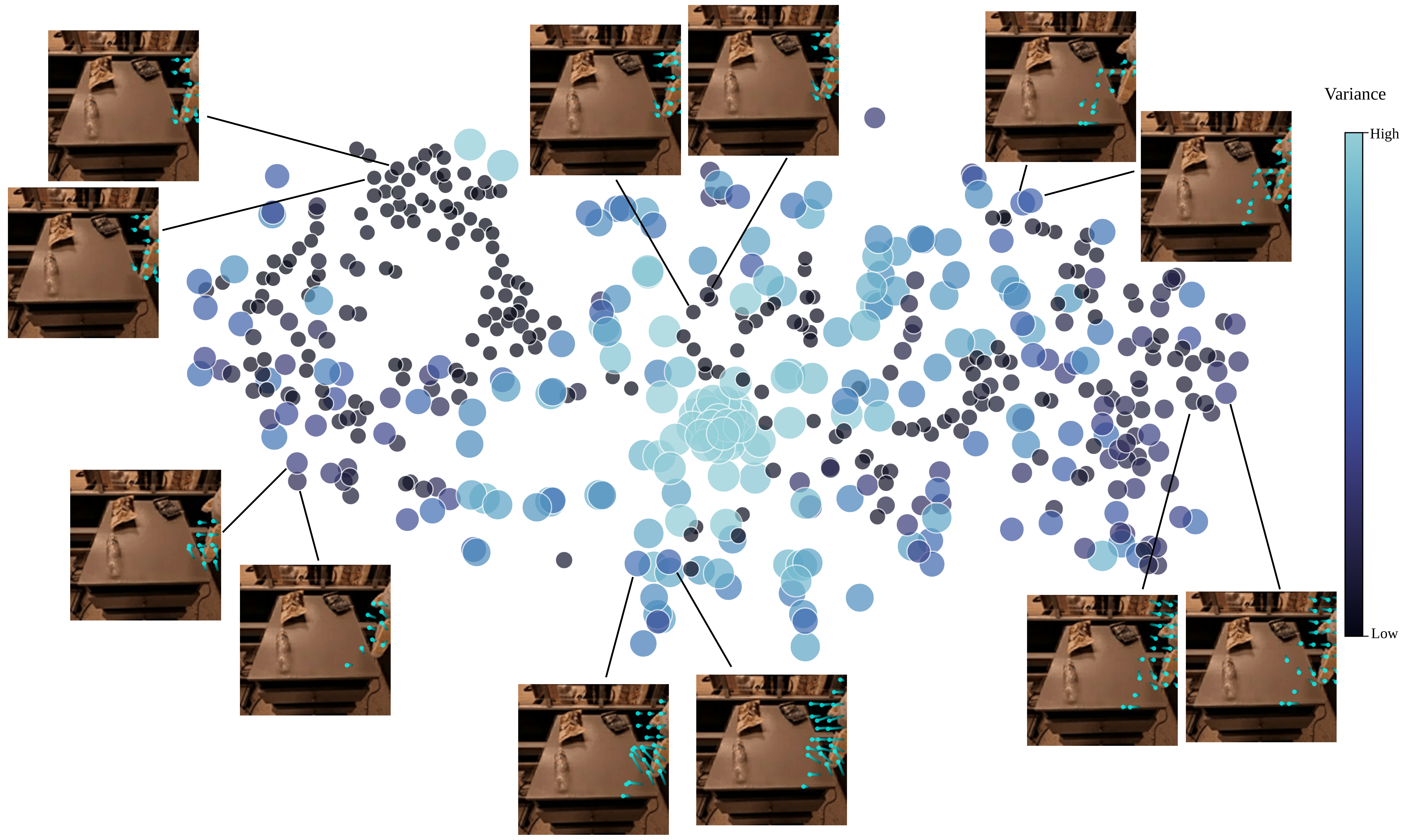}
    \caption{\textbf{Learned prior distributions visualization.} We plot the t-SNE of the learned means. The learned variances are represented by the color and size of the circles. We display the optical flow generated by the LAM decoder while reconstructing the same image using different latent actions.}
    \label{fig:priors}
\end{figure}

\begin{figure}[h!]
    \centering
    \includegraphics[width=\textwidth]{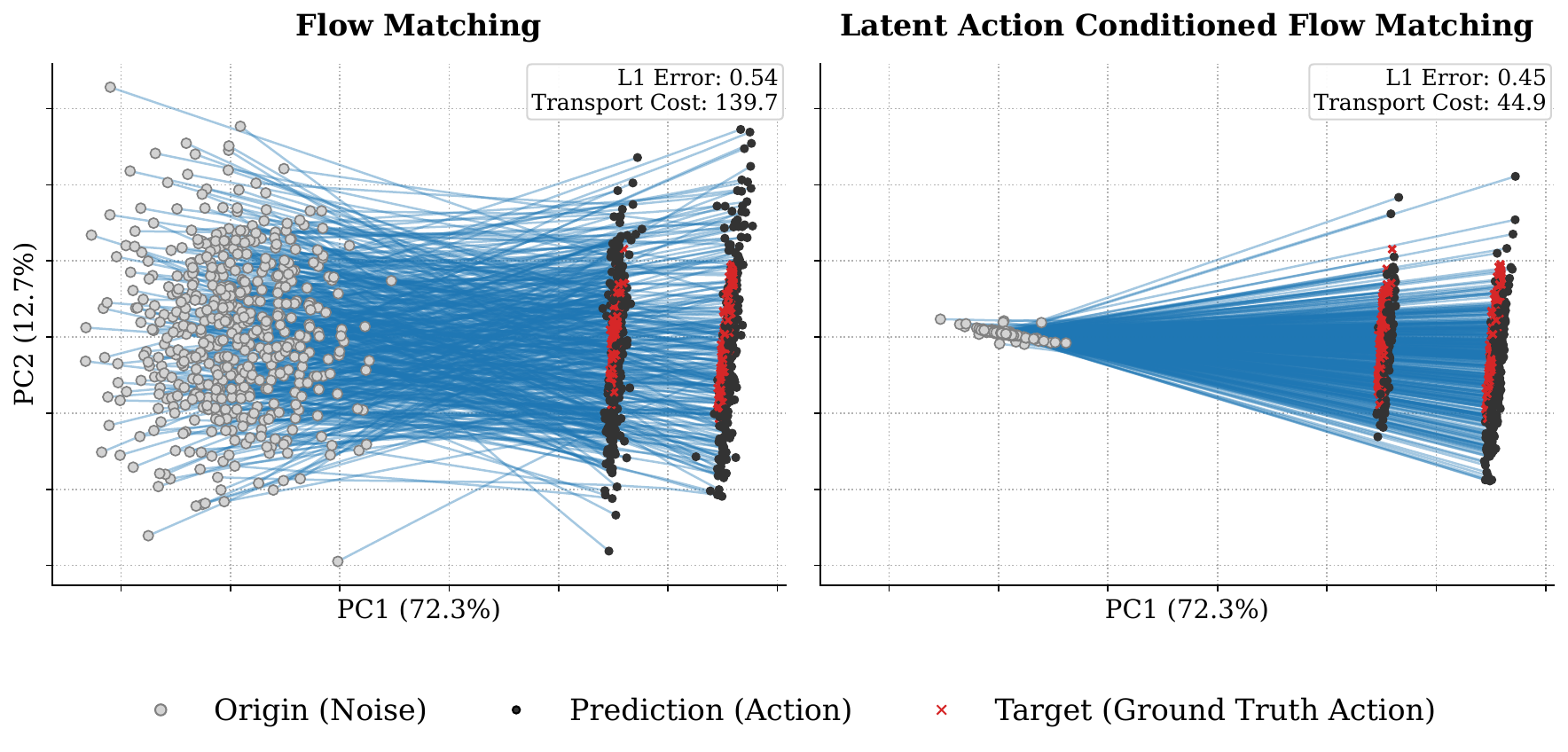}
    \caption{\textbf{Vector fields visualization.} We compare the denoising process of the standard and latent action guided flow matching policies. The gray dots are the sampled noise points, the black dots are the final predicted actions, and the blue lines connecting the dots represents the flow matching vector fields. The target actions are represented by the red crosses, being the same on both plots. We present the average L1 distance of the final predicted action to the target action and the average transport cost for the batch. The explained variance ratio is given for both principal components.}
    \label{fig:vector-fields}
\end{figure}

We provide Figures~\ref{fig:priors}~and~\ref{fig:vector-fields}, which allow us to better understand our method qualitatively. Figure~\ref{fig:priors} shows that latent actions that represent similar movements will have similar learned priors. Figure~\ref{fig:vector-fields} illustrates how LAFM reduces the initial displacement between source and target samples, producing shorter and less entangled transport trajectories.

\section{Latent action model}
\label{appendix:lam}

The latent action model (LAM) architecture pairs an inverse dynamics encoder—which compresses two or more consecutive frames into a latent action—with a forward dynamics decoder that uses this latent action alongside historical frames to predict future frames, as shown in Figure \ref{fig:method} (left). Crucially, to prevent the encoder from simply memorizing the target frames rather than extracting meaningful actions, an information bottleneck must be applied. This is typically enforced by discretizing the latent action space or by regularizing a continuous latent action space. Although the ideal aim of a latent action is to isolate and represent the behavior of specific agents, in practice, these models capture all visual dynamics within a scene, including extraneous factors like camera motion or background movement. Nonetheless, because the video dynamics in robotic manipulation datasets arise almost exclusively from the robot's movements, this is not necessarily a limitation for our use case.

\begin{table}[b]
    \caption{LAM hyperparameters.}
    \centering
    \setlength{\tabcolsep}{6pt}
    \renewcommand{\arraystretch}{1.2}
    \definecolor{grayrow}{gray}{0.90}
    \begin{tabular}{l|c}
        \toprule
        Model Dimension & 512 \\
        Feed-Forward Dimension & 2048 \\
        Encoder Layers & 8 \\
        Decoder Layers & 8 \\
        Attention Heads & 8 \\
        Dropout & 0.1 \\
        Image Size & 128 \\
        Patch Size & 16 \\
        \midrule
        Learning Rate & $5 \cdot 10^{-5}$ \\
        Batch Size & 128 \\
        Weight Decay & $10^{-4}$ \\
        Gradient Steps & 300,000 \\
        Optimizer & AdamW~\cite{adamw} \\
        Learning Rate Scheduler & Cosine Decay \\
        Warm-up Steps & - \\
        \bottomrule
    \end{tabular}
    \label{tab:hyper-param-lam}
\end{table}

Following \textcite{genie}, we implement a Spatial-Temporal Transformer~\cite{st} as our LAM encoder. The LAM decoder is implemented as a traditional Transformer. Since the decoder processes a single input frame, temporal attention layers are not required~\cite{univla}. To enforce the information bottleneck, the latent action space is discretized using NSVQ~\cite{nsvq}. All LAMs are trained with the configuration given in Table \ref{tab:hyper-param-lam}. Each training takes around 40 hours and is performed using a single NVIDIA H100 GPU. For all our experiments, with the exception of the ablations detailed in Appendix~\ref{appendix:lam-ablations}, we use the same LAM, which is trained in the Fractal~\cite{rt-1} dataset, using a codebook size of 512.

To ensure that latent actions capture coarse movements rather than low-level actions, we train the model using frame pairs sampled one second apart. For datasets with multiple cameras, a single camera view $\mathbf{I}^j$ is chosen and all latent actions are predicted using this view: $\ell_a \coloneqq \phi_e(\mathbf{I}_t^j, \mathbf{I}_{t+T}^j)$ where $T$ corresponds to the frequency of the dataset. Since all datasets used in our experiments are recorded at frequencies higher than 1 Hz, a single latent action effectively corresponds to a chunk of real actions. After training, the model is used to automatically annotate a target dataset with latent actions. LIBERO and the real-world dataset alike have two camera views, one external and another attached to the end-effector of the robot. For both cases, we extract the latent actions using the external view.

\section{Latent action model ablations}
\label{appendix:lam-ablations}

In this section, we analyze how different design choices in the LAM affect the performance of the final policy. All ablations are conducted on LIBERO-90. We keep the policy architecture and training configuration fixed across experiments, and only vary the LAM used to label the dataset.

\begin{figure}[h]
    \centering
    \includegraphics[width=\textwidth]{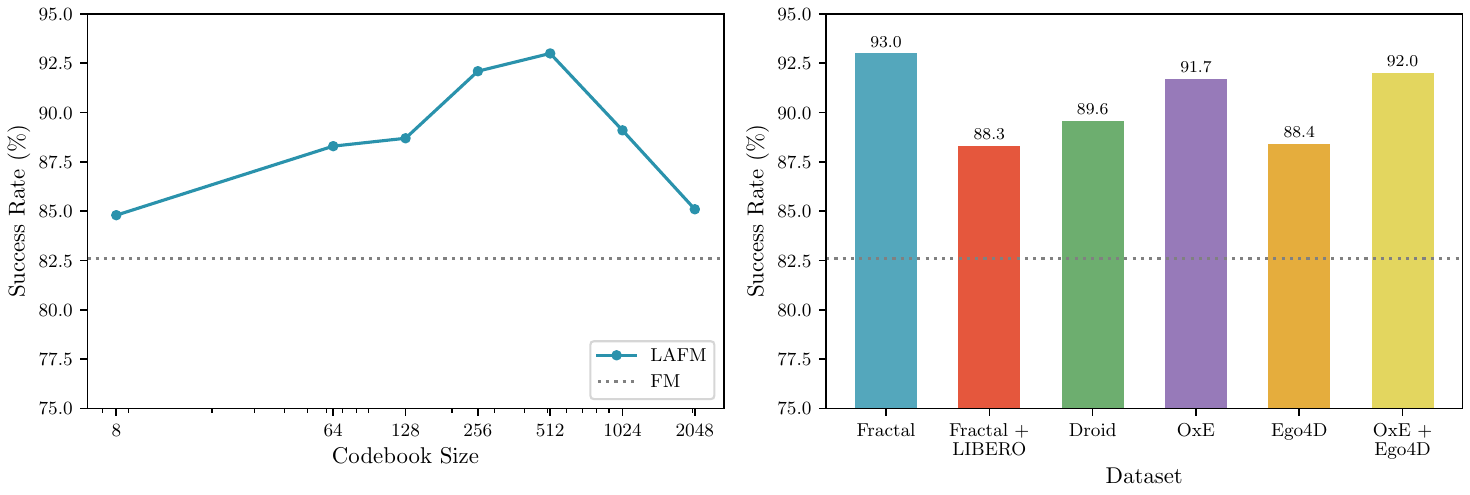}
    \caption{Impact of different LAM design choices on the final policy performance. We ablate the codebook size (left) and the dataset used during LAM training (right), and report the resulting success rate on LIBERO-90.}
    \label{fig:ablation}
\end{figure}

\subsection{Codebook size}

An important design choice is the size of the LAM codebook, as it directly determines the number of priors used in our method. Ideally, this size should match the number of motion primitives present in the target dataset. To study this effect, we train all LAMs on the Fractal dataset~\cite{rt-1} using the standard configuration described in Appendix~\ref{appendix:lam}, while varying the codebook size.

Figure~\ref{fig:ablation} shows the performance of the resulting policies. We observe that both excessively small and excessively large codebooks degrade performance, suggesting that under- and over-partitioning the latent action space are both harmful. Intuitively, these results are coherent with what we would expect. Not having enough priors to cover all motion primitives in the target dataset does not exploit the full potential of our method, while having too many priors leads to redundant distributions. Nevertheless, our method consistently outperforms standard flow matching across all prior counts.

\subsection{Training dataset}

We investigate how the choice of dataset used to train the LAM influences downstream performance. Our hypothesis is that, as long as the dataset contains sufficient motion diversity to capture the relevant motion primitives, its exact distribution should have limited impact.

To test this, we first consider single-source manipulation datasets: Fractal~\cite{rt-1}, which differs in embodiment from the target setting, and Droid~\cite{droid}, which shares the same embodiment but differs in visual distribution. LIBERO alone lacks sufficient diversity to train a meaningful LAM, the decoder can memorize future frames from the visual input alone, making the latent actions learned uninformative. We instead approximate an in-distribution setting by training on a mixture of Fractal and LIBERO.

We further evaluate a subset of OpenX Embodiment (OxE)~\cite{oxe}, using the sampling distribution proposed by \textcite{univla}, as well as Ego4D~\cite{ego4d}, a large-scale human manipulation video dataset. Finally, we consider a combination of OxE and Ego4D, where Ego4D constitutes 3.0\% of the mixture following \textcite{univla}.

The results in Figure~\ref{fig:ablation} show that adding Ego4D to the OxE mixture results in a small performance improvement. Nevertheless, using Fractal by itself outperforms both large dataset combinations. This suggests that quality is more important than quantity. Having a consistent dataset that covers the motion priors of interest seems to be more important than having a larger but less consistent data distribution. We posit that the lower performance when using Droid is due to the large number of different camera views, which makes it more challenging to extract consistent motion primitives. 

The lower performance on Ego4D is unsurprising, as it exhibits the largest distribution gap from our target datasets. Much of the apparent dynamics in Ego4D videos arises from camera motion, leading to learned priors that are not useful for our policy. Additionally, the inherent shakiness of egocentric footage introduces substantial visual noise, which complicates latent action learning. Finally, mixing Fractal with LIBERO results in the worst performance. Given the high predictability of the LIBERO data, even though mixing it with Fractal allows for stable LAM training, the resulting learned motion priors are worse, given that the model is able to predict the future frame regardless of the latent action for a part of its training data. 

\section{Policy training details and hyperparameters} \label{appendix:policies}

We give the implementation details for all our policy models and baselines. All models are trained in a single NVIDIA H100 GPU. Batch size is chosen to maximize GPU memory usage and the learning rate is adapted accordingly. Each training takes around 20 hours. All policies are trained and evaluated using the LeRobot library~\cite{lerobot}.

\subsection{LAFM and FM}

\begin{figure}[h!]
    \centering
    \includegraphics[width=0.7\linewidth]{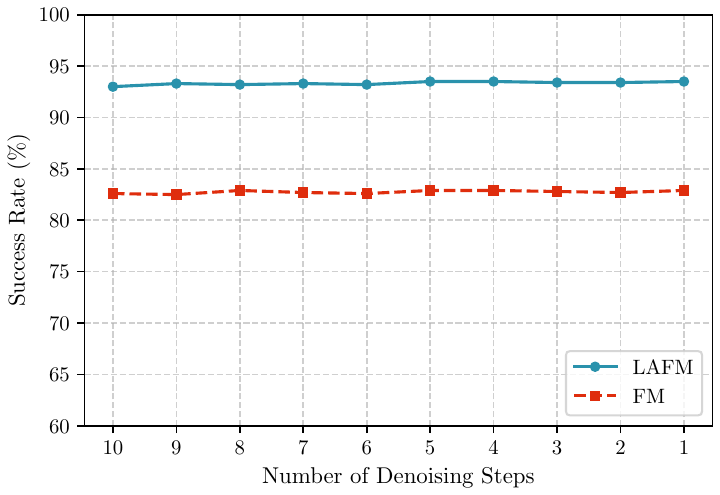}
    \caption{\textbf{Ablation on the number of denoising steps.} We plot the success rate on LIBERO-90 for LAFM and FM policies using different numbers of denoising steps during inference.}
    \label{fig:denoising-steps}
\end{figure}

For all experiments, we use the same configuration for LAFM and FM. We use ModernBERT \cite{modernbert} as frozen text encoder. To reduce the computational overhead, we implement a caching system, saving the embeddings of each task and re-using them when the tasks reappear. The image encoder is a ResNet18 \cite{resnet} pretrained on ImageNet-1k~\cite{imagenet}. We use a chunk size ($H$) of 10 actions for LIBERO and 20 actions for the real world experiments. We execute all 10 actions of the chunk when evaluating in LIBERO and execute 15 out of the 20 on real-world experiments. During inference, we use $\delta = 0.1$, performing 10 denoising steps for all experiments. Nevertheless, both LAFM and FM are quite robust to this hyperparameter, being able to maintain equivalent performance even with a single denoising step, as shown in Figure~\ref{fig:denoising-steps}. The other model and training hyperparameters are given in Table~\ref{tab:hyper-param-fm}.

\begin{table}[h!]
    \caption{LAFM and FM model and training hyperparameters.}
    \centering
    \setlength{\tabcolsep}{6pt}
    \renewcommand{\arraystretch}{1.2}
    \definecolor{grayrow}{gray}{0.90}
    \begin{tabular}{l|c}
        \toprule
        Model Dimension & 512 \\
        Feed-Forward Dimension & 3200 \\
        Encoder Layers & 8 \\
        DiT Layers & 8 \\
        Attention Heads & 8 \\
        Dropout & 0.1 \\
        \midrule
        Learning Rate & $8 \cdot 10^{-5}$ \\
        Batch Size & 448 \\
        Weight Decay & $10^{-2}$ \\
        Gradient Steps & 200,000 \\
        Optimizer & AdamW~\cite{adamw} \\
        Learning Rate Scheduler & Cosine Decay \\
        Warm-up Steps & 10,000 \\
        \bottomrule
    \end{tabular}
    \label{tab:hyper-param-fm}
\end{table}

\paragraph{Flow matching step sampling.}

During training, we follow \textcite{pi0} and sample the flow matching step $\tau$ from a scaled Beta distribution that prioritizes noisier samples and adds a cutoff value of 0.999. Specifically:
\[
u \sim \mathrm{Beta}(1, 1.5), \quad \tau = 0.999\,u, \quad \tau \in [0, 0.999]
\]

\paragraph{LAFM loss scalars.}

We set both $\lambda_1$ and $\lambda_2$ from Equation~\ref{eq:final_loss} to $10^{-3}$, which empirically seems to work the best in our experiments.

\subsection{ACT}

For ACT, we use the same configuration as in the original implementation \cite{act}. We use the same training and inference chunk sizes as for LAFM for all experiments. Table~\ref{tab:hyper-param-act} lists the remaining hyperparameters used.

\begin{table}[h!]
    \caption{ACT model and training hyperparameters.}
    \centering
    \setlength{\tabcolsep}{6pt}
    \renewcommand{\arraystretch}{1.2}
    \definecolor{grayrow}{gray}{0.90}
    \begin{tabular}{l|c}
        \toprule
        Model Dimension & 512 \\
        Feed-Forward Dimension & 3200 \\
        VAE Encoder Layers & 4 \\
        Policy Encoder Layers & 4 \\
        Policy Decoder Layers & 1 \\
        Attention Heads & 8 \\
        Dropout & 0.1 \\
        \midrule
        Learning Rate & $8 \cdot 10^{-5}$ \\
        Batch Size & 512 \\
        Weight Decay & $10^{-2}$ \\
        Gradient Steps & 200,000 \\
        Optimizer & AdamW~\cite{adamw} \\
        Learning Rate Scheduler & - \\
        Warm-up Steps & - \\
        \bottomrule
    \end{tabular}
    \label{tab:hyper-param-act}
\end{table}

 \subsection{${\pi_0}$}

 For $\pi_0$, we follow the exact same configuration as the original implementation \cite{pi0}, initializing the model from the released pre-trained weights. The original chunk size of 50 is kept during training, but we execute the same number of actions as for the LAFM experiments during inference. Training hyperparameters are shown in Table~\ref{tab:hyper-param-pi0}.

\begin{table}[h!]
    \caption{$\pi_0$ training hyperparameters.}
    \centering
    \setlength{\tabcolsep}{6pt}
    \renewcommand{\arraystretch}{1.2}
    \definecolor{grayrow}{gray}{0.90}
    \begin{tabular}{l|c}
        \toprule
        Learning Rate & $5 \cdot 10^{-5}$ \\
        Batch Size & 8 \\
        Gradient Steps & 200,000 \\
        Optimizer & AdamW~\cite{adamw} \\
        Learning Rate Scheduler & Cosine Decay \\
        Warm-up Steps & 1,000 \\
        \bottomrule
    \end{tabular}
    \label{tab:hyper-param-pi0}
\end{table}

Given the large scale of $\pi_0$, the maximum batch size allowed in a single GPU is much smaller than for the other polices studied. Nevertheless, we opt for a compute equivalent approach between all models, training them for the same number of gradient steps, which results in approximately the same number of GPU-hours used per training. Note that $\pi_0$ is being fine-tuned, while the other policies are being trained from scratch.

\section{Real-world experiments details} \label{appendix:experiments}

\begin{figure}[h]
    \centering
    \includegraphics[width=\linewidth]{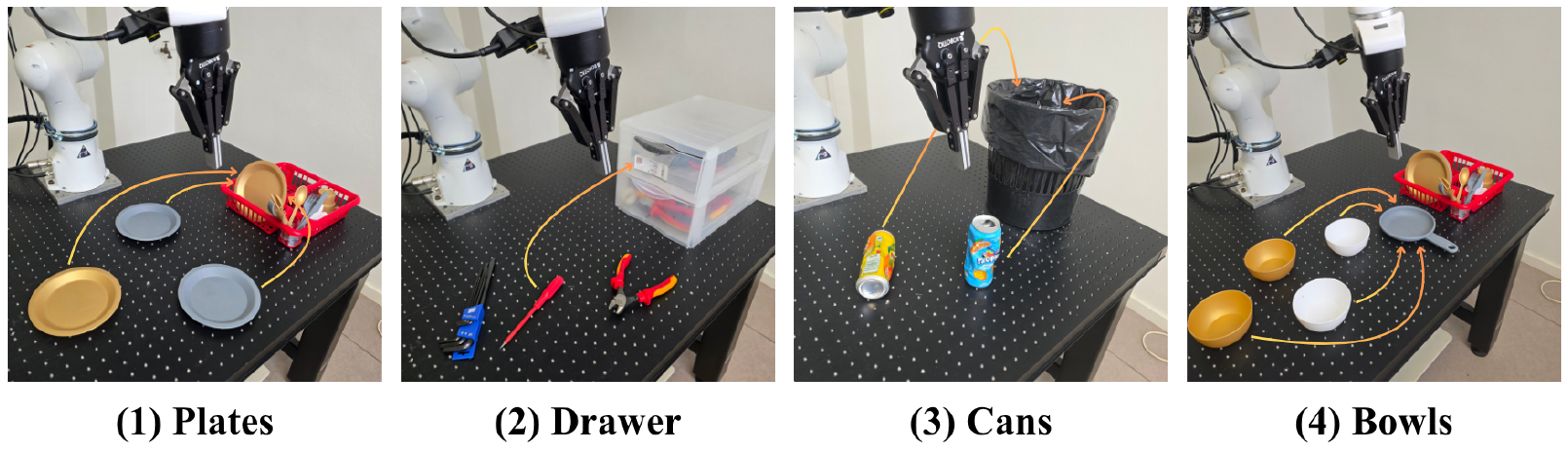}
    \caption{Real-world tasks examples.}
    \label{fig:tasks}
\end{figure}

Figure~\ref{fig:tasks} illustrates the four real-world tasks proposed for our experiments.

\subsection{Completion Score (CS) criteria}

For each task, we define a set of landmarks. The completion score is computed by adding the number of landmarks achieved during the episode and dividing it by the total number of landmarks in that same episode. 

\paragraph{Plates.} One to four plates are spread across the table. The position of each plate is chosen randomly. The dish dryer rack position does not change. The landmarks used to compute the completion score are: picking up a plate; placing a plate in the dish dryer rack. The score is therefore relative to the number of plates in a given scene. In order to compute a fair and comparable score across all models, we perform the same number of evaluation episodes per plate count with each model (e.g. five episodes with four plates, five episodes with three plates, and five episodes with two plates). We do the same for the cans and bowls task, which also have a variable number of possible objects in the scene.

\paragraph{Drawer.} The screwdriver is placed in a random position on the table. The drawer cabinet position is kept fixed. The completion score landmarks are: opening the drawer; picking up the screwdriver; placing the screwdriver inside the drawer; closing the drawer.

\paragraph{Cans.} One to two cans are placed in a random position on the table. The trash can position is kept fixed. The completion score landmarks are: picking up a can; throwing a picked can in the trash.

\paragraph{Bowls.} One to four bowls are placed randomly on the table. The position of the pan in which the bowls should be stacked is kept fixed. The completion score landmarks are: picking up a bowl; placing a bowl on top of the pan, for the first bowl, or placing a bowl on top of the bowl pile, for the others.

\section{Proof sketch for Proposition 1}
\label{appendix:proofs}

We compare the expected squared transport distance under two source distributions:
(i) a single isotropic Gaussian prior, and
(ii) a latent-conditioned prior aligned with action modes.

\paragraph{Setup.}
Let the conditional action distribution decompose as
\[
p(\mathbf{A}_t|\mathbf{o}_t)
=
\sum_{k=1}^{K}
\pi_k(\mathbf{o}_t)\, p_k(\mathbf{A}_t|\mathbf{o}_t),
\]
where each component $p_k(\mathbf{A}_t|\mathbf{o}_t)$ has mean $\mathbf{m}_k(\mathbf{o}_t)$ and covariance $\mathbf{S}_k$.

We analyze the expected squared distance $\mathbb{E}\|\mathbf{A}_t - \epsilon\|^2$ under the two constructions.

\paragraph{Case 1: Single isotropic prior.}

Let $\epsilon \sim \mathcal{N}(\mathbf{0}, \mathbf{I})$. Then:
\[
\mathbb{E}\|\mathbf{A}_t-\epsilon\|^2
=
\mathbb{E}\|\mathbf{A}_t\|^2
+
\mathbb{E}\|\epsilon\|^2
-
2\,\mathbb{E}[\mathbf{A}_t^\top \epsilon].
\]

Since $\epsilon$ is independent of $\mathbf{A}_t$ and has zero mean, the cross-term vanishes:
\[
\mathbb{E}[\mathbf{A}_t^\top \epsilon] = 0.
\]

Thus:
\[
\mathbb{E}\|\mathbf{A}_t - \epsilon\|^2
=
\mathbb{E}\|\mathbf{A}_t\|^2
+
\mathrm{Tr}(\mathbf{I}).
\]

Using the mixture decomposition:
\[
\mathbb{E}\|\mathbf{A}_t\|^2
=
\sum_{k} \pi_k(\mathbf{o}_t)
\left(
\|\mathbf{m}_k(\mathbf{o}_t)\|^2 + \mathrm{Tr}(\mathbf{S}_k)
\right).
\]

\paragraph{Case 2: Latent-conditioned prior.}

Let $\epsilon_k \sim \mathcal{N}(\boldsymbol{\mu}_k, \boldsymbol{\Sigma}_k)$ with $k \sim \pi(\mathbf{o}_t)$. Assume perfect mode coupling, i.e., the same latent index $k$ is used for both $\mathbf{A}_t$ and $\epsilon_k$. Then:
\[
\mathbb{E}\|\mathbf{A}_t - \epsilon_k\|^2
=
\sum_k \pi_k(\mathbf{o}_t)\,
\mathbb{E}_{\epsilon_k, \mathbf{A}_t \sim p_k}
\|\mathbf{A}_t - \epsilon_k\|^2.
\]

Expanding the squared norm:
\[
\mathbb{E}\|\mathbf{A}_t - \epsilon_k\|^2
=
\mathbb{E}\|\mathbf{A}_t - \boldsymbol{\mu}_k\|^2
+
\mathbb{E}\|\epsilon_k - \boldsymbol{\mu}_k\|^2
-
2\,\mathbb{E}[(\mathbf{A}_t - \boldsymbol{\mu}_k)^\top (\epsilon_k - \boldsymbol{\mu}_k)].
\]

Again, since $\epsilon_k$ is independent of $\mathbf{A}_t$ and centered at $\boldsymbol{\mu}_k$, the cross-term vanishes, yielding:
\[
\mathbb{E}\|\mathbf{A}_t - \epsilon_k\|^2
=
\mathbb{E}\|\mathbf{A}_t - \boldsymbol{\mu}_k\|^2
+
\mathrm{Tr}(\boldsymbol{\Sigma}_k).
\]

We further decompose:
\[
\mathbb{E}\|\mathbf{A}_t - \boldsymbol{\mu}_k\|^2
=
\|\mathbf{m}_k(\mathbf{o}_t) - \boldsymbol{\mu}_k\|^2
+
\mathrm{Tr}(\mathbf{S}_k).
\]

Thus:
\[
\mathbb{E}\|\mathbf{A}_t - \epsilon_k\|^2
=
\sum_k \pi_k(\mathbf{o}_t)
\left(
\|\mathbf{m}_k(\mathbf{o}_t) - \boldsymbol{\mu}_k\|^2
+
\mathrm{Tr}(\mathbf{S}_k)
+
\mathrm{Tr}(\boldsymbol{\Sigma}_k)
\right).
\]

\paragraph{Comparison.}

For the isotropic prior:
\[
\mathbb{E}\|\mathbf{A}_t - \epsilon\|^2
=
\sum_k \pi_k(\mathbf{o}_t)
\left(
\|\mathbf{m}_k(\mathbf{o}_t)\|^2
+
\mathrm{Tr}(\mathbf{S}_k)\right)
+
\mathrm{Tr}(\mathbf{I}).
\]

For the latent-conditioned prior:
\[
\mathbb{E}\|\mathbf{A}_t - \epsilon_k\|^2
=
\sum_k \pi_k(\mathbf{o}_t)
\left(
\|\mathbf{m}_k(\mathbf{o}_t) - \boldsymbol{\mu}_k\|^2
+
\mathrm{Tr}(\mathbf{S}_k)
+
\mathrm{Tr}(\boldsymbol{\Sigma}_k)
\right).
\]

If $\boldsymbol{\mu}_k \approx \mathbb{E}_{\mathbf{o}_t}[\mathbf{m}_k(\mathbf{o}_t)]$ and $\mathrm{Tr}(\boldsymbol{\Sigma}_k) \leq \mathrm{Tr}(\mathbf{I})$, then
\[
\mathbb{E}\|\mathbf{A}_t - \epsilon_k\|^2
\;\leq\;
\mathbb{E}\|\mathbf{A}_t - \epsilon\|^2,
\]
with strict inequality whenever the mode means are non-zero.

\paragraph{Interpretation.}

The isotropic prior requires transporting samples from a single global reference point to all action modes, while the latent-conditioned prior initializes samples closer to their corresponding modes. This reduces both the average displacement and the complexity of the learned transport field. This analysis focuses on squared Euclidean transport and provides an intuitive proxy for transport complexity; empirical results (Section \ref{sec:vector-vis}) corroborate that similar reductions occur in learned flow trajectories.

\paragraph{Remark on covariance assumption.}
If the action space is normalized to $[-1,1]^d$, then $\mathrm{Tr}(\boldsymbol{\Sigma}_k)\le d=\mathrm{Tr}(\mathbf{I})$ follows naturally, so the condition on the covariance is mild in practice.

\paragraph{Remark on perfect mode coupling assumption.}
The result assumes perfect mode coupling, i.e., that the latent index $k$ used to sample $\epsilon_k$ matches the component generating $\mathbf{A}_t$. In practice, this condition may not hold. However, as long as the correct mode is selected with high probability, the expected transport cost remains reduced, since most samples are initialized near their corresponding modes.

\section{Broader impacts statement} \label{appendix:impact}

Although the advancement of capable robotic manipulation through methods such as LAFM can enhance productivity and assist in hazardous tasks, it also carries the risk of labor displacement in sectors reliant on physical work. Ensuring a net positive impact requires a proactive focus on workforce retraining, education, and ensuring equitable access to these technologies to prevent exacerbating already excessive socioeconomic inequality. 

\end{document}